\documentclass{article} 

\usepackage{hyperref}
\usepackage{url}

\usepackage[utf8]{inputenc}
\usepackage[T1]{fontenc}
\usepackage{amsmath,amssymb,amsfonts}
\usepackage{booktabs}
\usepackage{authblk}
\usepackage{multirow}
\usepackage{colortbl}
\usepackage{hhline}
\usepackage{graphicx}
\usepackage{subcaption}
\usepackage{xcolor}
\usepackage{microtype}
\usepackage[numbers]{natbib} 

\newcommand{\model}{CoVisco}
\newcommand{\Tseg}{T_{\mathrm{seg}}}
\newcommand{\nframe}{n_{\mathrm{f}}}

\title{CoVisco: Codec-Native Vision Encoder with Native Token Compression for Unified Image-Video Understanding}

 \author[1,2]{Yulong Liu}
 \author[1]{Xiaotian Han\thanks{Project Lead}}
 \author[1]{Junyuan Shang}
 \author[1]{Yuchen Ding}
 \author[1]{Zhenyu Zhang}
 \author[1]{Shuohuan Wang}
 \author[3]{Guibo Zhu}
 \author[2]{Sirui Han\thanks{Corresponding Author}}
 \author[1]{Dianhai Yu}

\affil[1]{ERNIE Team, Baidu Inc.}
\affil[2]{ The Hong Kong University of Science and Technology}
 
\affil[3]{Institute of Automation, Chinese Academy of Sciences (CASIA)}

\begin{document}

\maketitle


 \begin{abstract}
Vision-language models face a fundamental scaling bottleneck: the number of visual
 tokens grows with both temporal duration and spatial resolution, making long-video
 understanding expensive for the vision encoder and the language model. Existing
 methods often compress visual tokens after dense encoding, creating a mismatch
 between the representation used during training and the compact interface required
 at deployment.
We present \model{}, a codec-native vision encoder with native token compression
for unified image-video understanding. By combining codec-native input support
with segmented attention, \model{} can encode long visual inputs in a single forward pass without forming dense patch-to-patch
interactions across all frames. Each temporal segment is equipped with learnable
\emph{abstract tokens} that
learn a compact segment-level representation, while fine-grained patch tokens
remain available throughout the encoder. Alternating intra-segment and
abstract-communication layers preserve video-level context through the
abstract-token channel. A lightweight selector further exposes either abstract
tokens alone or abstract tokens augmented with a runtime-selected subset of patch
tokens, yielding a compact visual interface that reduces the visual context and
prefill burden of downstream MLLMs while retaining fine-grained evidence when
needed.  Pretrained with contrastive objectives on 565M image--text
pairs and 6.4M videos, \model{} shows competitive performance on video-oriented
embedding and multimodal understanding benchmarks. In the evaluated
four-segment, 64-frame setting, abstract-only inference uses only 400 visual
tokens while achieving video-understanding performance close to, and on some
benchmarks exceeding, OneVision-Encoder. Selected patch tokens further improve
fine-grained video reasoning. Project URL:
\url{https://github.com/ernie-research/CoVisco.git}
 \end{abstract}

\section{Introduction}
\label{sec:intro}

Unified vision-language models must process images and videos through a common
visual interface, but video token counts grow with both duration and spatial
resolution. A 256-frame video at $448\times448$ already corresponds to roughly
$262\mathrm{K}$ patch tokens, creating a bottleneck for both visual encoding and
LLM prefill. The central challenge is therefore not only how to sample frames,
but how to learn a visual representation whose bandwidth can adapt to the
downstream task \citep{li2023blip2,qwen3vl}.

Most existing methods address this bottleneck after dense visual encoding, by
pruning, merging, or summarizing patch tokens before they reach the LLM
\citep{chen2024fastv,bolya2022tome,li2023blip2,li2023llama-vid}. Such post-hoc
compression can create a train--deploy mismatch: the encoder is trained with
dense patch flow, but the deployed interface must preserve information after
many tokens have been removed. This is especially problematic for video, where
brief actions, small objects, or text may occupy only a few patches in a few
frames. Encoder-internal compression improves this alignment, but methods that
retain only a compressed trace can permanently remove evidence needed for OCR,
spatial reasoning, or fine-grained temporal understanding.

Recent codec-native encoders use motion and residual signals to reduce the visual
input before or during encoding, including OneVision-Encoder
\citep{tang2026onevision} and Mage-VL \citep{magevl2026}. CoVisco combines this
codec-native input capability with a segmented vision architecture designed for
long visual inputs. Its segmented attention avoids dense patch-to-patch
interaction across the full temporal sequence, while abstract-mediated global
attention preserves video-level communication through a compact channel. As a
result, CoVisco can encode longer videos, including 256-frame inputs, in a single
forward pass and expose only a compact visual interface to the downstream MLLM,
reducing the visual context and prefill burden it must process. In codec mode,
codec-based selection first reduces the input, after which abstract tokens and
the selector determine which encoded tokens are exposed to the LLM. The same
abstract-token interface and segmented attention also handle uniformly sampled
and frame-collage inputs. Unlike methods that equate compression with permanent
deletion, CoVisco preserves fine-grained patch tokens inside the encoder and
exposes them on demand.

We present \model{} (\textbf{CO}dec-native \textbf{VIS}ion encoder with native
 token \textbf{CO}mpression), a unified image-video encoder built around three
ideas. First, each segment contains learnable abstract tokens that are directly
pooled by complementary image--text, image--image, and video--text objectives,
providing a compact segment representation. Second, alternating intra-segment
and abstract-communication layers restrict cross-segment interaction to the
abstract-token channel, enabling long visual inputs to be encoded without dense
patch-to-patch attention across all frames. Third, a lightweight selector exposes
abstract-only or abstract-plus-top-$K$ outputs to the LLM, providing a compact
visual interface that reduces the visual context and prefill burden of downstream
MLLMs while preserving fine-grained evidence for detail-sensitive tasks. The same
encoder supports both codec-native input and uniformly sampled or frame-collage
input.

Across the reported evaluations, \model{} provides a compact unified interface:
it supports codec-native and uniform visual inputs, encodes long videos with
segmented abstract-mediated attention, and exposes as few as 400 visual tokens
to the downstream LLM in the evaluated 64-frame setting. It is competitive with
selected larger models on video-oriented representation and understanding tasks,
while image and visual-document performance remains weaker, particularly for
OCR-heavy benchmarks. Our contributions are threefold:
\begin{enumerate}
  \item \textbf{Native compact representation with preserved fine-grained flow.}
  Abstract tokens learn a compact representation inside the ViT, while patch
tokens remain available for full-token or top-$K$ downstream use.
  \item \textbf{Abstract-mediated segmented video encoding.} Alternating
  attention layers restrict cross-segment interaction to abstract tokens,
  avoiding dense patch-to-patch attention while retaining a global communication
  path.
  \item \textbf{Dynamic unified image-video interface.} A selector supports
  abstract-only and abstract-plus-top-$K$ deployment across codec, uniform, and
  collage inputs, and a unified multi-resolution pretraining recipe transfers
  across retrieval and language-conditioned understanding tasks.
\end{enumerate}

\section{Related Work}
\label{sec:related}

\subsection{Vision encoders and codec-native video processing}
Contrastive vision encoders such as CLIP, SigLIP, SigLIP2, and MetaCLIP
\citep{radford2021clip,zhai2023siglip,tschannen2025siglip2,xu2023metaclip}
form the standard visual front-end for VLMs. Recent systems such as MoonViT
\citep{kimi_vl} support native-resolution inputs, while OneVision-Encoder
\citep{tang2026onevision} shows that codec motion and residual signals can
provide useful input sparsity. Long-video systems further combine frame
sampling, temporal aggregation, positional encoding, codec-aware selection, or
streaming memory. Sampling-based methods reduce frames or merge patches before
encoding; codec-structured methods such as CoViAR, EMA, OneVision-Encoder, and
Mage-VL exploit compressed-video structure; and streaming methods maintain a
compact temporal memory \citep{wu2018coviar,zhao2025ema,tang2026onevision,magevl2026,videollmonline,zhang2025flashvstream}.
These approaches expose a trade-off between temporal coverage, computational
cost, and fine-grained evidence. CoVisco supports both codec-native and uniform
frame inputs, while its segmented attention routes cross-segment communication
through abstract tokens for either input type.

\subsection{Visual token compression and compact interfaces}
Post-hoc methods prune, merge, or summarize tokens after visual encoding,
including importance-based dropping, token merging, and query-based
summarization \citep{rao2021dynamicvit,liu2023patchdropout,chen2024fastv,bolya2022tome,marin2023tokenpooling,li2023blip2,li2023llama-vid}. They are easy to attach to existing VLMs, but the dense encoder is not necessarily trained for the deployed compressed interface. Pretraining-time methods such as Video-LaVIT and OneVision-Encoder use codec-derived signals during visual encoding, while LLaVA-UHD v4 places early compression inside shallow ViT layers \citep{jin2024videolavit,tang2026onevision,fang2026llavauhdv4}. Learned query interfaces such as Perceiver, BLIP-2, and Chat-UniVi provide compact summaries, but typically replace or aggregate away fine-grained patch tokens \citep{jaegle2021perceiver,li2023blip2,lin2023chatunivi}. Recent studies also question whether attention importance reliably identifies redundancy and whether common benchmarks isolate compression sensitivity \citep{wen2025dart,wen2025token,liao2026vtc}. CoVisco instead supervises abstract tokens as a compact segment representation and the exclusive cross-segment communication channel, while preserving the fine-grained patch stream for full-token or top-$K$ use. This yields an adjustable interface that separates compression from permanent deletion.

\section{Method}
\label{sec:method}
\begin{figure*}[!t]
  \centering
  \resizebox{\textwidth}{!}{%
    \includegraphics{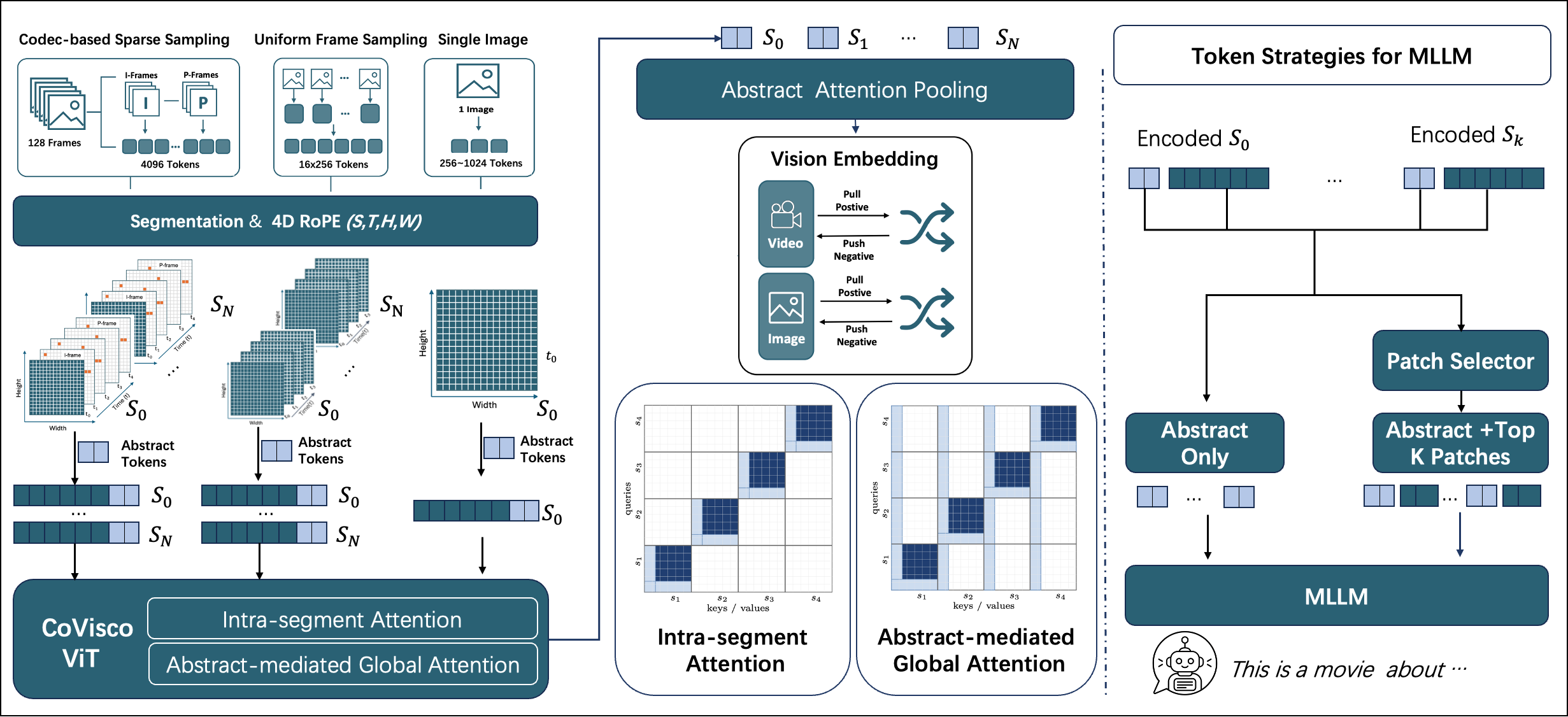}%
  }
  \caption{Framework of \model. (1)~The input video is divided into temporal
  segments and converted into visual tokens using codec-based, uniform,
  or collage sampling. (2)~Inside the ViT, alternating attention
  blocks perform full attention within each segment and abstract-mediated
  communication across segments. The central attention maps make this
  distinction explicit. Each segment contains learnable abstract tokens
  and fine-grained patch tokens. (3)~The ViT outputs both token types:
  abstract tokens are pooled and supervised by image--text, image--image, and
  video--text contrastive objectives, whereas fine-grained tokens are preserved
  throughout the encoder and remain available to a lightweight selector. At
  inference, the two outputs support abstract-only, abstract-plus-top-$K$, and
  full-token modes.}
  \label{fig:architecture}
\end{figure*}

We present \model{}, a codec-native vision encoder with a compact token
interface. The model interface, segmented attention, pretraining objectives,
token selector, and training curriculum are described below;
Figure~\ref{fig:architecture} summarizes the design.

\subsection{Problem setup and model interface}
\label{sec:prelim}
Given a video of $T$ frames at resolution $H \times W$, patchify with patch
size $p=14$ into $\nframe = \lceil H/p \rceil \lceil W/p \rceil$ tokens per
frame. The video is partitioned into $S$ temporal segments. During
pretraining, videos use four segments with a default temporal span of
$\Tseg=32$ frames per segment. At inference, the number of segments and the
number of frames assigned to each segment are runtime-configurable and need not
match the pretraining configuration; in particular, the per-segment frame
count may differ from 32, as reported in Section~\ref{sec:observations}.
For images, $S=1$. Each segment carries $Q$ learnable
\textbf{abstract tokens} ($Q=100$), so a segment with $\Tseg$ frames holds
$n = \nframe \Tseg + Q$ tokens in total. The encoder is a ViT-L backbone
(24 layers, hidden size $d=1024$, 16 heads,
$\sim$300M parameters) shared by images and videos; an image is simply the
degenerate case $S=1$.

The goal is twofold: (i) learn a compact representation---of size
$O(S \cdot Q)$---that is useful for retrieval and coarse understanding,
while preserving fine-grained tokens for optional downstream access; and (ii)
keep long-video computation tractable while avoiding dense cross-segment
patch-to-patch attention.

The encoder has three functional levels: patch tokens preserve local evidence,
abstract tokens summarize each segment, and abstract-mediated attention provides
video-level context. The selector exposes these levels through abstract-only,
abstract-plus-top-$K$, or full-token strategies.

The visual pipeline is
\emph{input} $\rightarrow$ \emph{patch embedding and segmentation}
$\rightarrow$ \emph{abstract-token augmented ViT}
$\rightarrow$ two output paths. The abstract path pools the final abstract-token
states for contrastive pretraining and compact language-model input, while the
patch path preserves the final patch-token states for optional selector-based
access. Given $S$ segments with $m$ visible patch tokens per segment,
patchification produces $\mathbf{X}\in\mathbb{R}^{S\times m\times d}$. A shared
learnable abstract-token matrix $\mathbf{Q}\in\mathbb{R}^{Q\times d}$ is expanded
over the segments and prepended to form
\begin{equation}
  \mathbf{H}^{(0)}_s = [\,\mathbf{Q}\,\Vert\,\mathbf{X}_s\,]
  \in\mathbb{R}^{(Q+m)\times d},
  \qquad s=1,\ldots,S.
\end{equation}
The resulting segment sequences pass through the alternating segmented ViT
layers described in Section~\ref{sec:attention}; 4D RoPE is applied to the
visible tokens using their segment, temporal, and spatial coordinates. At the
output, the sequence is split into abstract tokens
$\mathbf{A}\in\mathbb{R}^{S\times Q\times d}$ and fine-grained patch tokens
$\mathbf{P}\in\mathbb{R}^{S\times m\times d}$. The abstract path pools
$\mathbf{A}$ and applies the three contrastive projection heads described in
Section~\ref{sec:pretrain}, while the patch path preserves $\mathbf{P}$ for the
selector. For multimodal generation, the selected visual sequence is mapped by
a trainable MLP projector into the input space of the downstream language
model. Therefore, the language model can receive $\mathbf{A}$ alone,
$\mathbf{A}$ together with the selector's top-$K$ tokens from $\mathbf{P}$, or
all visual tokens, without changing the pretrained ViT.

\subsection{Abstract tokens as a compact interface}
\label{sec:abstract}
Each segment is prefixed with a shared set of $Q$ learnable parameters
$\mathbf{Q} \in \mathbb{R}^{Q \times d}$, so the input to the encoder is
$[\,\mathbf{Q} \,\Vert\, \mathbf{x}_1 \dots \mathbf{x}_{\nframe \Tseg}\,]$
per segment. Crucially, \textbf{only abstract tokens are directly pooled by the
contrastive heads} (Section~\ref{sec:pretrain}). The contrastive signal therefore
passes through $S \cdot Q$ output tokens, encouraging them---as a consequence of
the training objective---to become sufficiently informative for retrieval and
coarse understanding. We use ``native compression'' to refer to this learned,
compact output interface inside the vision encoder; it does not imply that all
fine-grained tokens are pruned during the ViT forward pass. Patch tokens are not
directly supervised by the contrastive heads, but they participate in every layer,
receive indirect gradient signals through the abstract representations, and remain
available for downstream use (Section~\ref{sec:selector}).

The design separates two roles: abstract tokens provide the compact supervised
representation, while patch tokens preserve fine-grained information for optional
downstream use.

\subsection{Segmented attention and abstract-mediated communication}
\label{sec:attention}
The 24 encoder layers alternate between two attention patterns.

Let $N$ denote the original number of patch tokens before adding abstract
tokens, let the $N$ patch tokens be divided evenly into $S$ segments, and let
$m = N/S$ be the number of patch tokens per segment. Each segment therefore
contains $n = m + Q$ tokens after adding its $Q$ abstract tokens.

\textbf{Intra-segment layers.} Tokens attend fully within their own segment;
the per-layer cost is $O(S(m+Q)^2)$. \textbf{Abstract-communication layers.}
Each query token attends to local patch tokens and the $SQ$ abstract tokens from
all segments, with cost $O(S(m+Q)(m+SQ))$. Thus, cross-segment information flows
exclusively through abstract tokens. Compared with dense attention over $N$
patch tokens, the leading patch-to-patch term changes from $O(N^2)$ to
$O(N^2/S)$, with additional abstract communication terms. This alternating
structure also defines the model's inductive bias: segment summaries retain
persistent semantics while fine-grained tokens preserve local evidence. Full
derivations are given in Appendix~\ref{sec:complexity-appendix}.

\subsection{Four-dimensional positional encoding}
\label{sec:rope}
We extend the 3D RoPE of prior codec-native encoders with a segment axis. The
rotary embedding factorizes head-dim frequencies into four groups with a
\textbf{2:4:5:5} split over (S, T, H, W): segments, being coarse, receive the
least positional bandwidth, while space receives the most. Patch positions are
computed on the \emph{full unpruned} spatio-temporal grid (a virtual grid of
$S \times (\Tseg{+}1) \times h \times w$) and then gathered for whichever
subset of patches is actually visible---so sparsity from codec sampling or
token selection does not change the underlying geometry, mirroring the shared
positional encoding principle of OneVision-Encoder \citep{tang2026onevision}.
Abstract tokens are assigned a reserved temporal index $t=0$ and a
$10\times10$ spatial layout ($Q=100$), providing a stable positional layout
distinct from the real frame tokens.
\subsection{Contrastive pretraining of CoVisco ViT}
\label{sec:pretrain}
\textbf{Objectives.} Pooled abstract tokens are passed to an attention-pooling
head and three projection heads trained with image--text, image--image, and
video--text SigLIP losses\cite{zhai2023siglip}. Because supervision reaches the encoder only through
abstract tokens, all three objectives shape the compact representation.

\textbf{Data and targets.} We use 480M LAION-2B image--caption pairs, 85M pairs
from \texttt{mvp-lab/LLaVA-OneVision-1.5-Mid-Training-85M}, and 6.4M 30/60-second
videos from \texttt{mvp-lab/LLaVA-OneVision-2-Data}. Target embeddings are
pre-extracted: SigLIP2 image/text targets are used for the LAION stream, and
Qwen3-VL-Embedding-8B text targets for the 85M image and video-caption streams.
The three heads map to the corresponding target spaces. The image and video
streams are mixed during training with per-stream batch sizes and gradient
accumulation.

\textbf{Training recipe.} Images use dynamic resolutions $\{224,336,448\}$;
videos use $224\times224$ inputs and four segments. Global batch sizes are 32K
for images and 3,200 for videos. Codec and uniform sampling are each used with
probability $1/2$; the uniform branch samples 16 frames and uses frame collage
with probability $1/2$. For the codec branch, we use input candidates derived
from H.265/HEVC (High Efficiency Video Coding).
Codec-derived visual candidates are selected independently within each segment
with a fixed per-segment token budget, and are then organized into the segment
structure before entering the ViT. This per-segment constraint is applied to
both training and evaluation data; the per-segment budget itself is not required
to be identical between training and inference.

\textbf{Positional augmentation.} During training, we randomly apply a global
offset to the segment IDs of both image and video inputs. This augmentation
exposes the ViT to a broader range of segment positions and helps it handle
inference inputs whose segment IDs extend beyond those used during training.
For single-image inputs, we likewise randomly shift the temporal IDs assigned to
the image patch tokens, while keeping the reserved temporal index of the
abstract tokens unchanged. This prevents the model from over-specializing to
the first-frame temporal position when processing images.

\subsection{Token selection and token strategies for MLLM}
\label{sec:selector}
While abstract tokens suffice for retrieval and coarse understanding, tasks
such as OCR, spatial grounding, and fine-grained temporal reasoning need
fine-grained evidence. \model{} therefore trains a lightweight \textbf{token
selector} that ranks the encoder's patch tokens for a given input, and
supports a continuum of deployment modes. In codec mode, these patch tokens
are already the subset retained by codec-based input selection; the selector
therefore performs a second, LLM-oriented compression over the codec-selected
stream. The selector follows the DynamicViT-style differentiable score-gating
idea for hard token selection \citep{rao2021dynamicvit}, while using a
segment-aware scoring transformer that is tailored to our abstract-token
interface. In our default configuration, the
scoring transformer contains two layers. Each layer first contextualizes patch
tokens within their own segment, then lets each patch token attend to the
abstract tokens of that segment, and finally applies a feed-forward network.
The resulting patch representations are mapped to scalar keep scores, and the
selector chooses top-$K$ patches independently for each segment. Appendix~\ref{sec:architecture-appendix}
gives the selector's tensor shapes and layer-level implementation details.
\begin{itemize}
  \item \textbf{Abstract-only}: feed the LLM the $S \cdot Q$ abstract tokens
  (exactly 400 tokens in the evaluated 64-frame, four-segment setting);
  \item \textbf{Abstract-plus-top-$K$}: augment the abstract tokens with the
  $K$ highest-ranked patch tokens from each segment, for a total of
  $S(Q+K)$ visual tokens. Here, $K$ is a runtime budget parameter that exposes a
  controllable accuracy--context trade-off;
  \item \textbf{Full}: expose all patch tokens when the visual context budget is
  unconstrained.
\end{itemize}
The selector is trained with Qwen3-1.7B on 1M images and 1M videos while the
ViT remains frozen; the selector, projector, and LLM are trainable. Images are
processed at their native resolutions, whereas videos use $224\times224$ inputs.
This preserves the pretrained visual backbone and lets the selector retain
fine-grained evidence when abstract tokens alone are insufficient.
\subsection{Training data and curriculum}
\label{sec:curriculum}
The entire training pipeline in this work consists of three stages,
using the following data and trainable modules: (1) contrastive
pretraining of the ViT on the image and video streams described above, followed
by freezing the ViT; (2) training the selector, projector, and Qwen3-1.7B on 1M
images and 1M videos with the ViT frozen, using native-resolution images and
$224\times224$ video inputs; and (3) instruction tuning of
Qwen3-4B-Instruct-2507 and the projector on 740K LLaVA-Next image instructions
and 800K LLaVA-Next-Videos samples, with the ViT and selector frozen. During
instruction tuning, images are also processed at their native resolutions,
whereas videos use $224\times224$ inputs.

\textbf{Compute.} ViT contrastive pretraining was conducted on 32 H800 GPUs. MLLM supervised fine-tuning (SFT) was conducted on 16 H800 GPUs.

\section{Experiments}
\label{sec:experiments}

\subsection{Evaluation as a unified embedding model}
We first ask whether \model{} learns a representation that remains useful when the
input modality and downstream objective change. This is a stronger test than
measuring only the quality of the visual features on the task used for
pretraining: a unified encoder should support image recognition, image--text
matching, video understanding, and visual-document retrieval through a common
embedding interface. We therefore evaluate \model{} in two complementary
settings: zero-shot image classification and image--text retrieval, followed by
MMEB-V2, which covers image, video, and visual-document embedding tasks.

\paragraph{Evaluation configuration.} In both settings, we use the visual
representation produced by the encoder, without an LLM or LLM-side token
selection. Zero-shot classification and image--text retrieval use the image
output of \model{} at the resolutions and sequence lengths listed in
Table~\ref{tab:zeroshot_reference}; the reference values are retained from the
corresponding SigLIP~2 comparison. MMEB-V2 instead evaluates the pooled
abstract-token representation with the benchmark's standard classification and
retrieval protocol. Thus, the MMEB-V2 results directly test whether the compact
representation learned inside the ViT retains task-relevant information across
modalities, rather than only whether it can serve as an input interface for a
language model.

\paragraph{Zero-shot image transfer.}
Table~\ref{tab:zeroshot_reference} reports zero-shot image classification and
image--text retrieval results. We retain the L/14 and L/16 blocks from the
SigLIP~2 comparison to contextualize the image-side transfer of \model{}, which
is trained as a unified image--video encoder with a compact abstract-token
interface.

\begin{table*}[!t]
  \centering
  \setlength{\tabcolsep}{0.52em}
  \renewcommand{\arraystretch}{1.1}
  \caption{\textbf{Zero-shot image classification and image-text retrieval reference.}
  Results are adapted from the SigLIP~2 zero-shot comparison~\citep{tschannen2025siglip2};
  only the ViT-L/14 and ViT-L/16 blocks are retained. Best values within each block are in bold formatting.}
  \label{tab:zeroshot_reference}
  \resizebox{\textwidth}{!}{%
  \begin{tabular}{lcclcccccccccc}
    \toprule
    &  &  &  & \multicolumn{4}{c}{ImageNet-1k} & \multicolumn{2}{c}{COCO} & \multicolumn{2}{c}{Flickr} & \multicolumn{2}{c}{XM3600} \\ \cmidrule(lr){5-8} \cmidrule(lr){9-10} \cmidrule(lr){11-12} \cmidrule(lr){13-14}
    ViT & Res. & Seq. & Model & val & v2 & ReaL & ObjNet & T$\rightarrow$I & I$\rightarrow$T & T$\rightarrow$I & I$\rightarrow$T & T$\rightarrow$I & I$\rightarrow$T \\
    \midrule
    \multirow[c]{6}{*}{L/14} & \multirow[c]{6}{*}{224} & \multirow[c]{6}{*}{256} & OpenCLIP~\citep{ilharco2021open} & 74.0 & 61.1 & -- & 66.4 & 46.1 & 62.1 & 75.0 & 88.7 & -- & -- \\
    &  &  & CLIP~\citep{radford2021clip} & 75.5 & 69.0 & -- & 69.9 & 36.5 & 56.3 & 65.2 & 85.2 & -- & -- \\
    &  &  & MetaCLIP~\citep{xu2023metaclip} & 79.2 & 72.6 & -- & 74.6 & {55.7} & -- & {83.3} & -- & -- & -- \\
    &  &  & CLIPA-v2~\citep{li2023clipav2} & 79.7 & 72.8 & -- & 71.1 & 46.3 & 64.1 & 73.0 & {89.1} & -- & -- \\
    &  &  & EVA-CLIP~\citep{sun2023eva} & {79.8} & {72.9} & -- & \textbf{75.3} & 47.5 & {63.7} & 77.3 & 89.7 & -- & -- \\
    &  &  & DFN~\citep{fang2024dfn} & \textbf{82.2} & \textbf{75.7} & -- & {74.8} & \textbf{59.6} & -- & \textbf{84.7} & -- & -- & -- \\
      \arrayrulecolor{lightgray}\hhline{|~|-------------|}
    \rowcolor{lightgray}
    &  &  & \model{} (ours) & 71.5 & 64.0 & 78.7 & 61.1 & 47.5 & 66.4 & 75.3 & 91.1 & \textbf{47.7} & \textbf{60.5} \\
    \arrayrulecolor{lightgray}\hhline{|~|-------------|}
    \rowcolor{lightgray}
     & 448&1024 &\model{} (ours)&73.3&66.1&80.3&65.1&48.5&\textbf{67.0}&77.3&\textbf{92.1}&47.4&60.2\\

    \arrayrulecolor{black}\hhline{|--------------|}
    \multirow[c]{5}{*}{L/16} & \multirow[c]{2}{*}{256} & \multirow[c]{2}{*}{256} & SigLIP~\citep{zhai2023siglip} & 80.5 & 74.2 & 85.9 & 77.9 & 51.2 & 69.6 & 81.3 & 92.0 & 30.9 & 40.1 \\
    &  &  & SigLIP2\citep{tschannen2025siglip2} & 82.5 & 76.8 & 87.3 & 83.0 & 54.7 & {71.5} & 84.1 & 94.5 & 46.5 & {56.5} \\
    \arrayrulecolor{lightgray}\hhline{|~|-------------|}
    & \multirow[c]{2}{*}{384} & \multirow[c]{2}{*}{576} & SigLIP~\citep{zhai2023siglip} & 82.1 & 75.9 & 87.1 & 80.9 & 52.8 & 70.5 & 82.6 & 92.9 & 31.4 & 39.7 \\
    &  &  & SigLIP2\citep{tschannen2025siglip2} & {83.1} & {77.4} & {87.6} & {84.4} & \textbf{55.3} & 71.4 & {85.0} & {95.2} & {47.1} & 56.3 \\
    \arrayrulecolor{lightgray}\hhline{|~|-------------|}
    & 512 & 1024 & SigLIP2\citep{tschannen2025siglip2} & \textbf{83.5} & \textbf{77.8} & \textbf{87.7} & \textbf{84.6} & {55.2} & \textbf{72.1} & \textbf{85.3} & \textbf{95.8} & \textbf{47.4} & \textbf{56.7} \\
    \arrayrulecolor{black}
    \bottomrule
  \end{tabular}%
  }
\end{table*}

Table~\ref{tab:zeroshot_reference} shows that \model{} preserves a usable
image-side representation while sharing the same encoder with videos. At
$224$ resolution, it obtains 71.5/64.0 on ImageNet-1k val/v2, 66.4 on COCO
image-to-text retrieval, and 91.1 on Flickr image-to-text retrieval; increasing
the resolution to $448$ improves these to 73.3/66.1, 67.0, and 92.1,
respectively. The improvement across both recognition and retrieval indicates
that the representation retains spatially useful information and is not tied to
a single fixed input grid. More importantly, these image results complement the
video results below: the compact interface learned by the unified encoder is
usable for both single-image semantics and cross-modal matching, rather than
being specialized only for temporal inputs. The gap to the strongest L/14 and
L/16 image encoders also makes clear that unification comes with a trade-off in
peak image-only accuracy, but the results establish a meaningful image-side
transfer capability for a model whose primary design goal is unified,
codec-native image--video encoding.

\paragraph{MMEB-V2 evaluation.}
Table~\ref{tab:mmeb-v2} evaluates the pooled abstract-token representation on
image, video, and visual-document embedding tasks using the benchmark's
standard classification and retrieval protocols. No language model or
LLM-side token selection is involved, allowing us to measure the transferability
of the compact representation itself.

\begin{table}[!t]
  \centering
  \renewcommand{\arraystretch}{1.2}
  \caption{
    Results on the MMEB-V2 benchmark~\citep{meng2025vlm2vec}. CLS:
    classification, RET: retrieval, VDR: ViDoRe, VR: VisRAG.
    $\textsuperscript{\dag}$: link to the model's homepage.
  }
  \label{tab:mmeb-v2}
  \resizebox{\textwidth}{!}{%
  \begin{tabular}{l c c c c c c c c}
    \toprule
    \multirow{2}{*}{\textbf{Model}}
    & \multirow{2}{*}{\textbf{Vision Tower}}
    & \multicolumn{1}{c}{\textbf{Image}}
    & \multicolumn{2}{c}{\textbf{Video}}
    & \multicolumn{3}{c}{\textbf{VisDoc}} \\
    \cmidrule(lr){3-3} \cmidrule(lr){4-5} \cmidrule(lr){6-8}
    & & \textbf{CLS}
    & \textbf{CLS} & \textbf{RET}
    & \textbf{VDRv1} & \textbf{VDRv2} & \textbf{VR} \\
    \midrule
    \textbf{\# of Datasets} $\rightarrow$ &
    & 10
    & 5 & 5
    & 10 & 4 & 6 \\

    \midrule
    VLM2Vec~\citep{jiang2025vlm2vec}
    & 2B &
    58.7 &
    33.4 & 20.6 &
    49.8 & 13.5 & 51.8 \\
    VLM2Vec-V2~\citep{meng2025vlm2vec}
    & 2B &
    62.9 &
    39.3 & 28.8 &
    75.5 & 44.9 & 79.4 \\
    GME~\citep{zhang2025bridging}
    & 2B  &
    54.4 &
    34.9 & 25.6 &
    86.1 & 54.0 & 82.5 \\
    Ops-MM-embedding-v1\href{https://huggingface.co/OpenSearch-AI/Ops-MM-embedding-v1-2B}{$\textsuperscript{\dag}$}
    & 2B &
    68.1 &
    53.6 & 41.8 &
    76.4 & 53.2 & 77.6 \\
    RzenEmbed~\citep{jian2025rzenembed}
    & 2B  &
    68.5 &
    50.4 & 46.6 &
    87.1 & 55.1 & 87.2 \\

    \cmidrule{1-8}
    VLM2Vec~\citep{jiang2025vlm2vec}
    & 8B &
    62.7 &
    39.1 & 29.0 &
    56.9 &  9.4 & 59.1 \\
    GME~\citep{zhang2025bridging}
    & 8B  &
    57.7 &
    37.4 & 28.4 &
    89.4 & 55.6 & 85.0 \\
    Ops-MM-embedding-v1\href{https://huggingface.co/OpenSearch-AI/Ops-MM-embedding-v1-7B}{$\textsuperscript{\dag}$}
    & 8B &
    69.7 &
    59.7 & 45.7 &
    80.1 & 59.6 & 79.3 \\
    RzenEmbed~\citep{jian2025rzenembed}
    & 8B  &
    70.6 &
    58.8 & 51.0 &
    89.7 & 60.7 & 88.7 \\

    \midrule
    IFM-TTE\href{https://interestfm-tte.github.io/}{$\textsuperscript{\dag}$}
    & 8B &
    76.7 &
    60.5 & 51.7 &
    85.2 & 71.5 & 92.7 \\
    Seed-1.6-embedding-0615\href{https://seed1-6-embedding.github.io/}{$\textsuperscript{\dag}$} & - &
    76.1 &
    55.0 & 51.3 &
    85.3 & 56.6 & 84.7 \\
    Seed-1.6-embedding-1215\href{https://seed1-6-embedding.github.io/}{$\textsuperscript{\dag}$} & - &
    75.0 &
    85.2 & 59.1 &
    90.0 & 60.3 & 90.0 \\

    \midrule
    \textbf{Qwen3-VL-Embedding-2B}
    & 2B &
    70.3 &
    71.9 & 53.9 &
    84.4 & 65.3 & 86.4 \\
    \textbf{Qwen3-VL-Embedding-8B}
    & 8B &
    74.2 &
    78.4 & 58.7 &
    87.2 & 69.9 & 88.7 \\

    \midrule
    \textbf{\model{} (ours)}
    &0.3B &
    55.1 & 65.1 &
    46.2 & 61.8 & 42.6&63.8 \\
    \bottomrule
  \end{tabular}%
  }
\end{table}

Despite using only a 0.3B vision tower, \model{} transfers across all three
benchmark families through the same abstract-token space. Its clearest strength
is video: it obtains 65.1 on video classification and 46.2 on video retrieval,
exceeding the listed 2B and 8B open-source baselines on video classification
except the Qwen3-VL embedding models, while remaining close to larger models on
video retrieval. The representation therefore preserves temporal information
that is useful for both discrimination and cross-modal matching, rather than
serving only as a compression target for the downstream VLM.

The transfer is broad but not uniform. Image classification reaches 55.1, while
visual-document scores are 61.8/42.6/63.8 on ViDoRe-v1, ViDoRe-v2, and VisRAG,
respectively, below the strongest larger baselines, particularly on
OCR- and document-centric tasks. We thus view MMEB-V2 as evidence that the
learned representation is useful across different tasks and modalities, not as
a claim of state-of-the-art performance on every benchmark: with a 0.3B vision
tower, one shared compact representation supports image recognition, video
understanding, video retrieval, and document retrieval, with especially strong
transfer to video.

\subsection{LMM Probing Evaluation}
\textbf{Evaluation configuration.} All multimodal understanding experiments use
Qwen3-4B-Instruct-2507 as the language backbone. We compare codec-guided
and uniformly sampled frame inputs. For \model{}, the LLM-side visual-token
budget is $S\cdot Q$ for abstract-only inference and $S(Q+K)$ for
abstract-plus-top-$K$ inference, where $K$ is the number of selected patch
tokens per segment. This budget is distinct from the input budget processed by
the ViT. In the video benchmarks, codec inputs use the original H.265/HEVC
stream without re-encoding; uniform inputs use 8 sampled frames for the
reference models. The matched OneVision-Encoder codec setting uses 10{,}368
codec-selected tokens from 64 frames, while its dense frame setting uses 8 frames at
$504\times504$; SigLIP2 uses 8 frames at $512\times512$. CoVisco codec inputs
are evaluated at $224\times224$ and $504\times504$, with 10{,}368 codec-selected
ViT input tokens, and the dense 64-frame condition is reported separately. For
image benchmarks, all images preserve their native resolutions. Token budgets
refer to visual context or encoder input as specified.

Table~\ref{tab:video_multimodal_benchmarks} and
Table~\ref{tab:image_multimodal_benchmarks} compare vision encoders under a
unified multimodal setting with Qwen3-4B-Instruct-2507 as the language backbone.

\begin{table*}[t]
  \centering
  \setlength{\tabcolsep}{4pt}
  \renewcommand{\arraystretch}{1.1}
  \caption{\textbf{Video benchmark comparison under a fixed language backbone.}
  All models are evaluated using Qwen3-4B-Instruct-2507. OneVision-Encoder
  uses a 10{,}368-token visual budget in both its codec and uniform-frame
  settings; these visual tokens enter its vision encoder and are forwarded to
  the language model. In the codec setting, the tokens are selected from a
  64-frame input using codec scores, whereas the frame setting uses 8 uniformly
  sampled frames at $504\times504$. SigLIP2 uses 8 uniformly sampled frames at
  $512\times512$. For \model{}, \textbf{Codec} denotes codec-guided input
  selection and \textbf{Frame} denotes uniform frame sampling. The notation is
  (sampling mode, input resolution, fine-grained token retention ratio for LLM); ``--''
  indicates that the corresponding setting was not evaluated.
  Bold values indicate the best performance among all rows shown for each
  benchmark.}
  \label{tab:video_multimodal_benchmarks}
  \resizebox{\textwidth}{!}{%
  \begin{tabular}{lccccccc}
    \toprule
    \textbf{Model / setting} & \textbf{MVBench} & \textbf{MLVU-dev} &
    \textbf{NExT-QA (MC)} & \textbf{VideoMME} & \textbf{Perception Test} &
    \textbf{TOMATO} & \textbf{LongVideoBench-Val-V} \\
    \midrule
    OV-Encoder (Codec,504,1.0)\cite{tang2026onevision} & {{52.4}} & 46.3 & {\textbf{75.6}} & {{53.4}} & {\textbf{60.3}} & 22.2 & {{50.4}} \\
    OV-Encoder-Frame (Frame,504,1.0)\cite{tang2026onevision} & 49.8 & {{49.4}} & 71.9 & 49.3 & 56.7 & 21.8 & 45.5 \\
    SigLIP2 (Frame,512,1.0)\cite{tschannen2025siglip2} & 47.2 & 48.4 & 70.6 & 46.8 & 56.0 & {{22.3}} & 45.2 \\
     \arrayrulecolor{lightgray}\hhline{--------|}
    \model{} (Codec,224,0.0) & 55.8 & 57.8 & 72.8 & 52.9 & 58.7 & 25.1 & 48.0 \\
    \model{} (Codec,224,0.4) & 56.7 & 58.5 & 74.0 & 55.7 & 59.7 & 26.0 & 50.5 \\
    \arrayrulecolor{lightgray}\hhline{--------|}
    \model{} (Codec,504,0.0) & 53.2 & 56.7 & 69.9 & 51.3 & 56.1 & 25.1 & 47.3 \\
    \model{} (Codec,504,0.4) & 53.2 & 56.7 & 71.3 & 53.0 & 57.0 & \textbf{26.4} & 48.2 \\
     \arrayrulecolor{lightgray}\hhline{--------|}
    \model{} (Frame,224,0.0) & 56.0 & \textbf{61.2} & 73.5 & 53.5 & 59.0 & 25.5 & 47.3 \\
    \model{} (Frame,224,0.2) & \textbf{56.9} & 59.8 & 73.9 & \textbf{57.0} & 59.5 & 25.7 & \textbf{52.7} \\
        \model{} (Frame,224,0.4) & 56.4 & 59.6 & 73.5 & \textbf{57.0} & 59.0 & 24.9 & 52.2 \\
    \bottomrule
  \end{tabular}}
\end{table*}

\begin{table*}[t]
  \centering
  \setlength{\tabcolsep}{4pt}
  \renewcommand{\arraystretch}{1.1}
  \caption{\textbf{Image benchmark comparison under a fixed language backbone.}
  All models are evaluated using Qwen3-4B-Instruct-2507. \textbf{Codec} denotes
  codec-guided input selection, whereas \textbf{Frame} denotes uniform frame
  sampling. Each setting is written as (sampling mode, input resolution,
  fine-grained token retention ratio for LLM); ``--'' indicates that the corresponding
  setting was not evaluated. Bold values indicate the best performance among
  all rows shown for each benchmark.}
  \label{tab:image_multimodal_benchmarks}
  \resizebox{\textwidth}{!}{%
  \begin{tabular}{lccccccccc}
    \toprule
    \textbf{Model / setting} & \textbf{AI2D} & \textbf{ChartQA} & \textbf{DocVQA} &
    \textbf{InfoVQA} & \textbf{MMBench-EN} & \textbf{OCRBench} &
    \textbf{OCRBench v2} & \textbf{MMStar} & \textbf{RealWorldQA} \\
    \midrule
    OV-Encoder (Codec,native,1.0)\cite{tang2026onevision} & 75.7 & 76.5 & 78.4 & 43.1 & 77.2 & 605 & {\textbf{26.3}} & 52.1 & 60.8 \\
    OV-Encoder-Frame (Frame,native,1.0)\cite{tang2026onevision} & 76.5 & {\textbf{77.8}} & {\textbf{79.5}} & {\textbf{45.5}} & 78.5 & {\textbf{630}} & 26.1 & 54.3 & 61.2 \\
    SigLIP2 (Frame,native,1.0)\cite{tschannen2025siglip2} & {\textbf{78.6}} & 76.4 & 75.0 & 42.0 & {\textbf{79.6}} & 621 & 26.1 & {\textbf{55.0}} & {\textbf{62.1}} \\
     \arrayrulecolor{lightgray}\hhline{|----------|}
    \model{} (Frame,native,0.0) & 66.8 & 17.4 & 20.5 & 20.7 & 68.2 & 261 & 21.2 & 43.8 & 45.5 \\
    \model{} (Frame,native,0.2) & 67.9 & 51.8 & 64.0 & 32.1 & 70.9 & 474 & 24.0 & 45.7 & 52.8 \\
    \model{} (Frame,native,0.8) & 69.8 & 68.8 & 75.1 & 42.2 & 71.8 & 567 & 24.0 & 48.8 & 54.1 \\
    \model{} (Frame,native,1.0) & 70.5 & 70.7 & 75.5 & 44.1 & 71.6 & 576 & 24.4 & 48.3 & 53.3 \\
    \bottomrule
  \end{tabular}}
\end{table*}

The multimodal results show that the value of the compact interface depends on
the granularity of evidence required by the task. On video benchmarks, the
abstract-only setting already exposes just 400 visual tokens in the four-segment
configuration, yet remains competitive with references that forward 10{,}368
visual tokens to the language model. This indicates that the abstract tokens
capture sufficient segment-level and temporal context for many video questions.
Adding a small number of selected patches provides a consistent gain on the
more detail-sensitive video metrics: for example, in the $224$-resolution frame
setting, increasing the fine-grained retention ratio from $0.0$ to $0.2$ improves
VideoMME from 53.5 to 57.0 and LongVideoBench-Val-V from 47.3 to 52.7. The
same pattern is visible in the codec setting, where the $0.4$ configuration
improves VideoMME from 52.9 to 55.7 and LongVideoBench-Val-V from 48.0 to 50.5.
Thus, the results support a division of labor: abstract tokens provide a compact
global summary, while selected patches restore localized evidence when the
question depends on a small object, text region, or brief event.

The comparison also separates two forms of selection that should not be
conflated. Codec-guided selection reduces the visual input processed by the
encoder, whereas the learned selector controls the visual context forwarded to
the language model. The dense 64-frame condition further shows that the
segmented architecture can retain useful performance when temporal coverage is
not reduced at the input. In contrast, the image results improve sharply as
more fine-grained tokens are exposed: for example, DocVQA rises from 20.5 in
abstract-only mode to 75.1 at a retention ratio of $0.8$. Nevertheless, the
image-centric scores remain below the strongest baselines, especially on
OCR-heavy tasks, suggesting that compact segment summaries alone are not a
substitute for high-resolution text evidence and OCR-oriented pretraining.

\begin{figure*}[t]
  \centering
  \includegraphics[width=\textwidth]{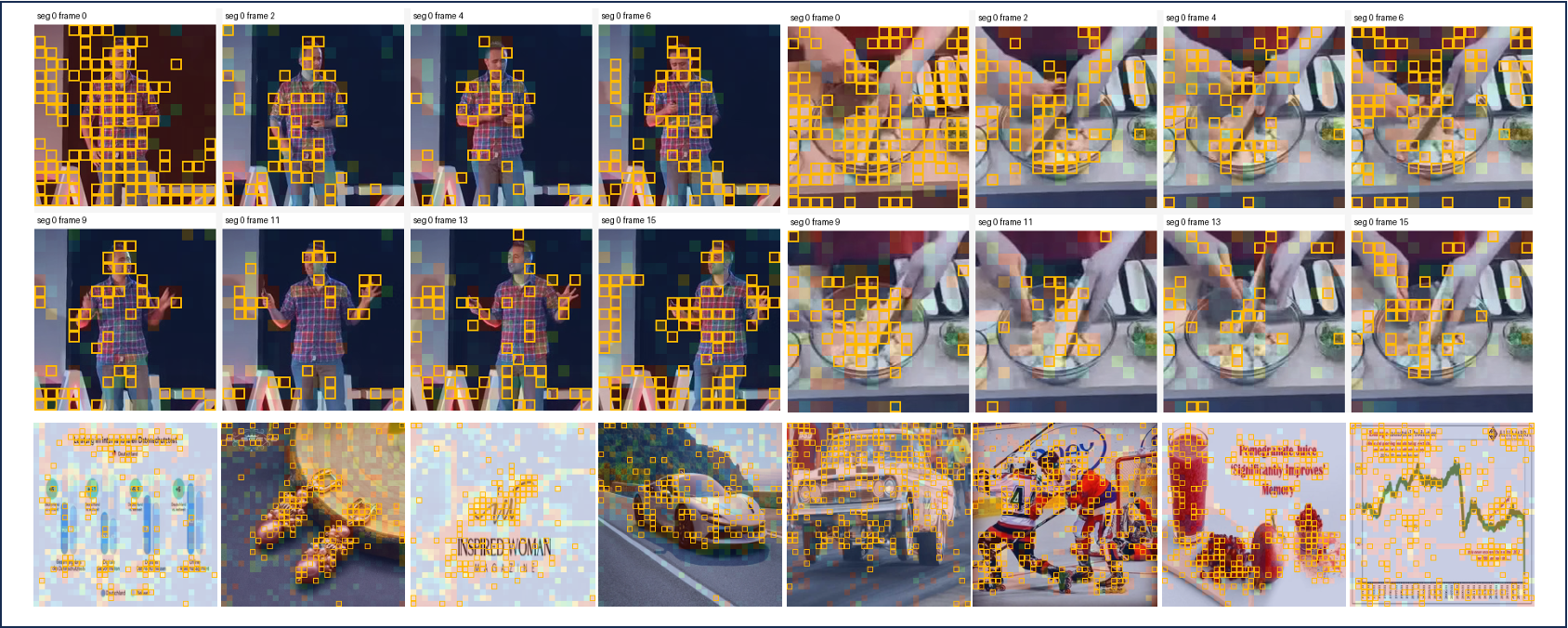}
  \caption{Qualitative visualization of the token selector on images and
  videos. Yellow boxes mark fine-grained patch tokens retained by the selector.
  The selected regions typically focus on objects, text, and localized changes,
  illustrating how selected patch tokens complement the abstract-token
  summaries.}
  \label{fig:selector_visualization}
\end{figure*}

These quantitative trends motivate a closer look at what the learned selector
actually retains. If the selected patches are complementary to the abstract
tokens, they should concentrate on localized regions that are difficult to
represent with a segment-level summary, rather than simply reproducing the
same global content. Figure~\ref{fig:selector_visualization} provides this
qualitative diagnostic. It is not intended as evidence of selector optimality,
but illustrates that the retained patches commonly cover objects, text, and
localized temporal changes, offering an interpretable counterpart to the
accuracy gains observed above.

\subsection{Long video understanding with different segment configuration}
\label{sec:observations}
\textbf{Evaluation configuration.} We study whether the compact
abstract-token interface remains reliable when the temporal granularity changes
at inference time. All experiments use Qwen3-4B-Instruct-2507 and the
abstract-only interface with $Q=100$ abstract tokens per segment and
$224\times224$ visual inputs. The language model is SFT-tuned with four
segments, but evaluation changes the number of sampled frames (64, 128, or
256), the number of segments ($S\in\{4,8,16\}$), and the sampling mode
(uniform frame or codec-guided input) without retraining. Consequently, the
LLM receives $S\times Q$ abstract tokens (400, 800, or 1{,}600 tokens), while
the encoder must accommodate different temporal lengths and segment layouts.
These experiments use the long-video setting described here and should not be
confused with the 10{,}368-token codec-matched input budget used in the
controlled encoder comparison.

Table~\ref{tab:segment-generalization} shows that the central behavior is
stable under changes in both video length and segmentation. With uniform frame
sampling, MLVU varies only from 61.2 to 62.6 across 64--256 frames and from
61.2 to 62.3 when the 128-frame input is repartitioned from 8 to 16 segments.
On LongVideoBench-Val, the same changes do not cause degradation: accuracy
increases from 47.3 at 64 frames and four segments to 49.9--50.3 at 128--256
frames, with only a 0.3-point difference between 8 and 16 segments at 128
frames. Thus, increasing temporal coverage or changing the segment granularity
does not require retuning the language model, even though the resulting
abstract-token budget changes by up to four times.

This robustness is consistent with the architecture introduced in
Sections~\ref{sec:method} and~\ref{sec:complexity-appendix}. Intra-segment
attention models local temporal and spatial evidence, while abstract tokens
provide the compact global communication path across segments. Changing $S$
therefore changes where local computation is factorized, but does not remove
the mechanism that carries video-level context. The results provide empirical
evidence that the abstract tokens are not tied to the four-segment SFT layout;
the pretrained encoder can preserve a useful global representation when a
longer video is split more finely and processed in one forward pass.

The codec rows further show that this stability is compatible with
codec-guided input, although its effect is benchmark-dependent. On
LongVideoBench-Val, codec sampling improves over uniform sampling at both 64
frames (48.0 vs.~47.3) and 256 frames (52.4 vs.~50.3), indicating that the
abstract interface can exploit codec-selected visual evidence. MLVU is more
sensitive to this sampling choice (57.8 at 64 frames and 59.9 at 256 frames),
so the result should not be read as uniform superiority across all tasks.
Rather, the consistent finding is that the segmented encoder remains usable
across frame counts, segment counts, and input modes, with the largest
variation attributable to the task-specific sampling policy rather than to a
failure to generalize across segment configurations.

\begin{table*}[t]
  \centering
  \caption{Long-video understanding under different segment configurations.
  The language model is SFT-tuned with four segments; evaluation changes the
  number of segments at inference time without retraining. All configurations use
  the abstract-only mode. The abstract-token budget is $S\times Q$ with $Q=100$.}
  \label{tab:segment-generalization}
  \small
  \resizebox{\textwidth}{!}{
  \begin{tabular}{lcccccc}
    \toprule
    Benchmark & Resolution & Frames & Segments & Sampling & Token strategy & Accuracy (\%) \\
    \midrule
    MLVU & 224 & 64  & 4  & Frame & Abstract-only & 61.2 \\
     MLVU & 224 & 64  & 4  & Codec & Abstract-only & 57.8 \\
    MLVU & 224 & 128 & 8  & Frame & Abstract-only & 62.6 \\
    MLVU & 224 & 128 & 16 & Frame & Abstract-only & 62.3 \\
    MLVU & 224 & 256 & 16 & Frame & Abstract-only & 61.3 \\
    MLVU & 224 & 256 & 16 & Codec & Abstract-only & 59.9 \\
    \midrule
    LongVideoBench-Val & 224 & 64  & 4  & Frame & Abstract-only & 47.3 \\
    LongVideoBench-Val & 224 & 64  & 4  & Codec & Abstract-only & 48.0 \\
    LongVideoBench-Val & 224 & 128 & 8  & Frame & Abstract-only & 49.9 \\
    LongVideoBench-Val & 224 & 128 & 16 & Frame & Abstract-only & 50.2 \\
    LongVideoBench-Val & 224 & 256 & 16 & Frame & Abstract-only & 50.3 \\
    LongVideoBench-Val & 224 & 256 & 16 & Codec & Abstract-only & 52.4 \\
    \bottomrule
  \end{tabular}}
\end{table*}

\section{Conclusion}
\label{sec:conclusion}
We introduced \model{}, a unified image-video encoder with a native compact token
interface. Abstract tokens learn segment-level representations inside the ViT,
while fine-grained patch tokens remain available for downstream selection.
Segmented attention routes cross-segment communication through abstract tokens,
and deployment can vary from abstract-only to abstract-plus-top-$K$ inputs.
Across the reported evaluations, this design provides a compact alternative to
exposing all visual patches to the language model, with the clearest benefits on
video-oriented tasks: abstract-only inference provides a compact segment-level
representation, while selected patch tokens can restore local evidence when
fine-grained reasoning is required. The stable performance across different
frame counts, segment configurations, and sampling modes further indicates that
the abstract-token interface is not tied to a single temporal layout or
inference budget.

The current results also identify several limitations. The model does not use
OCR-specific pretraining, its instruction-tuning scale is modest, and the
abstract-token capacity is fixed at $Q=100$. Therefore, the OCR
results should be interpreted as transfer measurements, and the most effective
token budget remains task-dependent. Future work will improve OCR-oriented
transfer and recognition performance, study the capacity--accuracy trade-off,
and characterize end-to-end budget policies and efficiency more systematically.
Moreover, the compact segment-level interface of \model{} makes it naturally
suited to streaming video understanding, where visual information must be
encoded and updated incrementally under tight computation and latency budgets.
Our current design provides a useful foundation for more tightly unified
audio-visual encoding and for rethinking streaming video understanding around
compact, incrementally processed representations. Developing these directions
is an important avenue for future work.

\bibliography{iclr2027_conference}
\bibliographystyle{iclr2027_conference}

\appendix
\section{Detailed model architecture}
\label{sec:architecture-appendix}
This appendix makes the tensor organization and output interfaces explicit. The
same encoder is used for images and videos. An image is represented by one
segment ($S=1$), whereas a video is divided into $S$ temporal segments. Let
$m$ be the number of visible patch tokens in one segment, $Q$ the number of
abstract tokens, and $d=1024$ the hidden dimension. After patch embedding and
sampling, the visual input is represented as
\begin{equation}
  \mathbf{X} = [\mathbf{X}_1,\ldots,\mathbf{X}_S],
  \qquad \mathbf{X}_s\in\mathbb{R}^{m\times d}.
\end{equation}
The abstract tokens are a single shared parameter matrix, rather than a
separate parameter set for every segment:
\begin{equation}
  \mathbf{Q}\in\mathbb{R}^{Q\times d},
  \qquad
  \mathbf{H}^{(0)}_s = [\,\mathbf{Q}\,\Vert\,\mathbf{X}_s\,]
  \in\mathbb{R}^{(Q+m)\times d}.
\end{equation}
The matrix $\mathbf{Q}$ is expanded along the segment and batch dimensions at
runtime. Thus, the segments have independent token positions at the input but
share the same abstract-token parameterization. With the default $Q=100$, the
compact representation exposed by abstract-only inference contains $S Q$
tokens.

\subsection{Attention routing and positional inputs}
The ViT contains 24 alternating transformer layers. In an intra-segment layer,
the attention mask is block diagonal over segments: all tokens in segment $s$
act as queries and can read the $m+Q$ tokens in that segment only. In an
abstract-communication layer, all tokens in segment $s$ again act as queries;
each query can read its $m$ local patch tokens and the $S Q$ abstract tokens from
all segments. Writing $\mathcal{P}_s$ for the patch-token indices of segment $s$,
$\mathcal{A}_s$ for its abstract-token indices, and
$\mathcal{A}=\bigcup_{r=1}^{S}\mathcal{A}_r$ for all abstract-token indices,
the query and visible key sets are
\begin{equation}
  \mathcal{Q}_s^{\mathrm{intra}} = \mathcal{Q}_s^{\mathrm{comm}}
  = \mathcal{P}_s\cup\mathcal{A}_s,
  \qquad
  \mathcal{K}_{s}^{\mathrm{intra}} = \mathcal{P}_s\cup\mathcal{A}_s,
  \qquad
  \mathcal{K}_{s}^{\mathrm{comm}} = \mathcal{P}_s\cup\mathcal{A}.
\end{equation}

\begin{center}
\fbox{%
\begin{minipage}{0.94\linewidth}
\small
\textbf{Algorithm: Segmented attention with abstract-mediated communication.}
Given segmented hidden states $\{\mathbf{H}_s\}_{s=1}^{S}$, let
$\mathcal{P}_s$ and $\mathcal{A}_s$ denote the patch-token and abstract-token
indices in segment $s$, respectively. Define the query set for every segment as
$\mathcal{Q}_s=\mathcal{P}_s\cup\mathcal{A}_s$ and let
$\mathcal{A}=\bigcup_{r=1}^{S}\mathcal{A}_r$.

\textbf{For} each encoder layer $l=1,\ldots,24$:
\begin{enumerate}
  \item If layer $l$ is an intra-segment layer, use
  $\mathcal{Q}_s=\mathcal{P}_s\cup\mathcal{A}_s$ and set
  $\mathcal{K}_s \leftarrow \mathcal{P}_s \cup \mathcal{A}_s$ for every
  segment $s$. Every query attends only to tokens in its own segment.
  \item If layer $l$ is an abstract-communication layer, use the same query
  set $\mathcal{Q}_s=\mathcal{P}_s\cup\mathcal{A}_s$ and set
  $\mathcal{K}_s \leftarrow \mathcal{P}_s \cup \mathcal{A}$ for every segment
  $s$. Each query can attend to local patch tokens and all segment-level
  abstract tokens, but never to patch tokens from another segment.
  \item Apply self-attention from queries $\mathcal{Q}_s$ to keys/values in
  $\mathcal{K}_s$, followed by the transformer feed-forward block.
\end{enumerate}
After the final layer, split each segment's hidden states into abstract tokens
$\mathbf{A}$ and fine-grained patch tokens $\mathbf{P}$. Return $\mathbf{A}$ and
$\mathbf{P}$ as the two downstream output paths.
\end{minipage}%
}
\end{center}

This routing allows every segment to write local evidence into its abstract
 tokens and allows all segments to exchange information through the abstract
 channel, while disallowing direct cross-segment patch-to-patch attention. The
 4D rotary embedding receives the segment, temporal, height, and width
 coordinates of each visible token. Patch coordinates are gathered from the
 full virtual grid after sampling, while abstract tokens use the reserved
 summary positions specified in Section~\ref{sec:rope}.

\subsection{Output paths and downstream interfaces}
After the final transformer layer, the hidden states are split into
\begin{equation}
  \mathbf{A}\in\mathbb{R}^{S\times Q\times d},
  \qquad
  \mathbf{P}\in\mathbb{R}^{S\times m\times d},
\end{equation}
where $\mathbf{A}$ contains the abstract tokens and $\mathbf{P}$ contains the
fine-grained patch tokens. Only $\mathbf{A}$ is sent to the attention-pooling
head during contrastive pretraining. The pooled representation is then mapped
by three separate fully connected heads to the image--image
($1536$-dimensional), image--text ($1536$-dimensional), and video--text
($4096$-dimensional) target spaces. Each head has its own learnable logit scale
and bias and is trained with its corresponding SigLIP sigmoid loss. The
image--video-caption stream uses the video--text projection space, so it does
not introduce a fourth projection head.

The patch output $\mathbf{P}$ is not pooled away. During the selector stage, a
lightweight selector ranks these tokens and exposes a runtime-selected subset
of them. The selected visual tokens, either $\mathbf{A}$ alone or
$\mathbf{A}$ concatenated with selected tokens from $\mathbf{P}$, are passed
through the MLP projector used by the multimodal language model. The ViT is
frozen during this alignment stage; the selector, projector, and language
model are trained according to the curriculum in Section~\ref{sec:curriculum}.
At inference, changing $K$ changes only the number of fine-grained tokens
exposed to the language model and does not alter the encoder computation or
its pretrained parameters.

\subsection{Token selector details}
The selector receives the final ViT representations with the segment dimension
preserved. For a batch of inputs, its inputs are
\begin{equation}
  \mathbf{A}\in\mathbb{R}^{B\times S\times Q\times d},
  \qquad
  \mathbf{P}\in\mathbb{R}^{B\times S\times P\times d},
\end{equation}
where $\mathbf{A}$ denotes the per-segment abstract/query tokens and
$\mathbf{P}$ denotes the corresponding fine-grained patch tokens. Selection is
performed independently within each segment: the $(B,S)$ dimensions are
flattened temporarily, so a segment never competes directly with patches from
another segment.

The default selector is a two-layer scoring transformer. For each layer
$l$, let $\mathbf{Z}^{(l)}_s$ denote the patch-token features for segment $s$,
and let $\mathbf{A}_s$ denote its abstract tokens. The layer applies
\begin{align}
  \widetilde{\mathbf{Z}}^{(l)}_s
    &= \operatorname{LN}\!\left(\mathbf{Z}^{(l)}_s +
       \operatorname{SelfAttn}(\mathbf{Z}^{(l)}_s)\right), \\
  \widehat{\mathbf{Z}}^{(l)}_s
    &= \operatorname{LN}\!\left(\widetilde{\mathbf{Z}}^{(l)}_s +
       \operatorname{CrossAttn}(\widetilde{\mathbf{Z}}^{(l)}_s,
                                  \mathbf{A}_s,\mathbf{A}_s)\right), \\
  \mathbf{Z}^{(l+1)}_s
    &= \operatorname{LN}\!\left(\widehat{\mathbf{Z}}^{(l)}_s +
       \operatorname{FFN}(\widehat{\mathbf{Z}}^{(l)}_s)\right).
\end{align}
Here patch tokens are the queries in the cross-attention, while the
abstract/query tokens provide keys and values. The default implementation uses
8 attention heads, a $4d$ feed-forward hidden dimension when not otherwise
specified, GELU activation, and zero dropout. The layer count remains a
configuration parameter, with two layers used as the default selector
configuration.

After the second layer, a final LayerNorm and a linear score head produce one
logit per patch token:
\begin{equation}
  \ell_{s,i} = \operatorname{Linear}\!\left(
    \operatorname{LN}(\mathbf{Z}^{(2)}_{s,i})\right),
  \qquad
  p_{s,i}=\sigma(\ell_{s,i}).
\end{equation}
The default sigmoid score treats each patch as having an independent keep
probability. Invalid padded positions are assigned $-\infty$ logits and cannot
be selected. For each segment, the selector then chooses the $K$ patches with
largest scores, returning their indices, the full score map, and the selected
features. In the implementation, the scores are computed from the
contextualized selector features, while the selected feature values are
gathered from the corresponding original ViT patch tokens.

Hard top-$K$ selection is non-differentiable. Following the score-gating idea
of DynamicViT~\citep{rao2021dynamicvit}, the selected features are multiplied
by their continuous scores during training, which provides a straight-through
gradient path to the scoring network while retaining hard top-$K$ indices in
the forward pass. An optional Gumbel mode perturbs the logits during training
for stochastic exploration; evaluation uses deterministic scores. The selector
therefore returns
$\mathbf{P}_{\mathrm{sel}}\in\mathbb{R}^{B\times S\times K\times d}$ and
indices with shape $(B,S,K)$, which are concatenated with
$\mathbf{A}$ in the abstract-plus-top-$K$ deployment mode and then passed to
the MLP projector.

\section{Attention-cost analysis details}
\label{sec:complexity-appendix}
\textbf{Complexity scope.} The attention-cost expressions below characterize
encoder-side attention operations. They should be read separately from the
LLM-side token budgets reported in the main experiments and do not constitute a
wall-clock latency measurement. Let $N$ be the number of patch tokens before
adding abstract tokens, $S$ the number of temporal segments, $m=N/S$ the number
of patch tokens per segment, and $Q$ the number of abstract tokens prepended to
each segment. Each segment therefore has $n=m+Q$ tokens.

A dense ViT layer over the original patch sequence costs $O(N^2)$ attention
operations per layer. If the same dense attention were applied after adding all
$S Q$ abstract tokens, the cost would become
\begin{equation}
  O\!\left((N + SQ)^2\right)
  = O\!\left(N^2 + 2NSQ + S^2Q^2\right).
\end{equation}
This gives every patch token direct access to every other patch token, but it is
precisely the pattern that becomes prohibitive for long videos.

\model{} alternates two cheaper patterns. In intra-segment layers, each segment
attends only within its own length-$n$ sequence, giving
\begin{equation}
  O(S n^2)
  = O\!\left(S(m+Q)^2\right)
  = O\!\left(\frac{N^2}{S} + 2NQ + SQ^2\right).
\end{equation}
In abstract-communication layers, each token attends to its local segment patch
tokens and to the abstract tokens from all segments. Each of the $S(m+Q)$ query
tokens therefore attends to $m+SQ$ keys, giving
\begin{equation}
  O\!\left(S(m+Q)(m+SQ)\right)
  = O\!\left(\frac{N^2}{S} + NSQ + NQ + S^2Q^2\right).
\end{equation}
The leading dense patch-to-patch term is reduced from $N^2$ to $N^2/S$, while
all global communication is restricted to the abstract-token channel. When
segment length is fixed and longer videos are handled by increasing $S$, the
per-segment patch count $m$ and abstract count $Q$ remain constant. The segmented
design removes dense all-to-all patch interaction across segments, but the
abstract-communication layers still incur additional costs because each segment
attends to the abstract tokens of all segments. Thus, the method trades dense
cross-segment patch attention for local segment computation plus a compact global
communication channel. This design avoids forming an all-to-all patch sequence as
video duration increases, while its actual scaling depends on the number of
segments, the per-segment token count, and the abstract-token budget.

The analysis isolates attention operations and does not include
implementation-dependent constants such as FlashAttention kernel efficiency,
memory layout, or
data-loading overhead. Those constants affect wall-clock latency and should
therefore be evaluated separately from the operation-count comparison above.
\section{H.265/HEVC codec-native input construction}
\label{sec:h265-hevc}
The codec-based experiments use visual candidates derived from a video encoded
with H.265/HEVC (High Efficiency Video Coding). H.265/HEVC is a block-based,
inter-frame video coding standard that represents a video using intra-coded
reference frames and inter-coded frames. Inter-coded frames are reconstructed
from motion-compensated predictions plus residual signals, while intra-coded
frames provide independently decodable spatial references. These motion and
residual signals expose where visual content changes or where prediction is
less reliable, providing a codec-native basis for constructing a sparse visual
input. This appendix describes how we use that information; it does not treat
H.265/HEVC itself as a learned visual encoder.

The codec-native pipeline operates directly on
codec-aligned H.265/HEVC visual data without re-encoding the source videos. For
evaluation, we use the original compressed-video stream and its existing
group-of-pictures (GOP) structure. The reported temporal windows can contain 64,
128, or 256 frames; changing the window length changes the amount of video
processed by CoVisco, not the underlying video encoding or GOP structure.

For the codec-selected input, the budget entering the ViT and the budget
exposed to the LLM are explicitly tracked. For each segment $s$, codec
selection forms
\begin{equation}
  \mathcal{V}_{\mathrm{codec},s}
  = \operatorname{TopK}(\mathcal{V}_s; K_{\mathrm{codec}}),
  \qquad s=1,\ldots,S,
\end{equation}
where $K_{\mathrm{codec}}$ is fixed within each segment configuration and the
total number of codec-selected tokens entering the ViT is $S K_{\mathrm{codec}}$.
The codec-selection rule enforces equal token counts across temporal segments
within each input, while allowing the per-segment budget to vary between
training and inference configurations. The learned selector is applied after
ViT encoding and independently selects fine-grained patch tokens within each
segment for the LLM. Therefore,
 codec-selected tokens entering the ViT
  $\neq$tokens selected by the learned selector for the LLM
 
The two selection stages are complementary: the first controls the sparse
H.265/HEVC-native input, and the second controls the visual context exposed to
the language model. For OneVision-Encoder in the codec setting, its
10{,}368 codec-selected tokens serve both roles: they enter the vision encoder
and are forwarded as the visual tokens to the language model.

For comparison with OneVision-Encoder, the codec-matched CoVisco settings use
$S K_{\mathrm{codec}}=10{,}368$ visual tokens entering the ViT at both
the $224\times224$ and $504\times504$ input resolutions. The per-segment budget
is kept uniform across segments within each configuration, but may be changed
between training and inference. OneVision-Encoder uses the same 10{,}368-token
budget in its codec setting, selecting tokens from a 64-frame video according to
codec scores. The segment-count study additionally reports 64-, 128-, and
256-frame inputs, allowing us to evaluate whether the same H.265/HEVC-based input
pipeline and segmented visual encoder remain stable as the temporal window grows. The uniform-frame references instead use 8 densely
sampled frames, with $504\times504$ inputs for OneVision-Encoder and
$512\times512$ inputs for SigLIP2. These settings match the visual-token budget
where applicable while preserving the respective spatial resolutions and input
policies.

In addition to codec-matched inputs, CoVisco is evaluated with dense 64-, 128-,
and 256-frame inputs in the reported segment-count study. These experiments
retain the sampled frame tokens entering the segmented ViT rather than applying
the 10{,}368-token codec budget, and test whether the segmented attention design
can encode longer dense visual sequences without forming dense cross-segment
patch-to-patch attention. Uniform-frame and frame-collage inputs use the same
downstream abstract-token and learned-selector interfaces, but do not use
H.265/HEVC-derived sparse candidate selection.
\section{Controlled evaluation protocol}
\label{sec:controlled-pipeline}
The controlled comparison separates the contribution of the visual encoder from
that of the language backbone and instruction data. Each path uses the same
Qwen3-4B-Instruct-2507 language model, the same MLP projector family, and the
same instruction-tuning data described in Section~\ref{sec:curriculum}. The
paths differ only in the visual encoder and the visual-token interface exposed
to the language model. This design is intended to answer a narrow question: for
a fixed decoder and fixed multimodal training recipe, how much performance is
retained when dense frame tokens are replaced by a compact abstract-token
interface plus an optional selected patch budget?

Path A uses \model{} with segmented abstract tokens and, where specified,
abstract-plus-top-$K$ fine-grained tokens. Path B replaces the visual encoder with
a codec-native OneVision-Encoder-style baseline under matched sampling and token
budgets. Path C uses a frame-based SigLIP2-style encoder under uniform sampling.
For fair comparison, the protocol reports both budget levels when they differ:
CoVisco's codec-selected input budget counts tokens entering the vision encoder,
whereas its abstract-only or abstract-plus-top-$K$ budget counts visual tokens
forwarded to the language model. For OneVision-Encoder in the codec setting,
the 10{,}368 codec-selected tokens are both the input to the vision encoder and
the visual tokens forwarded to the language model. Codec and frame variants are
reported separately because they make different assumptions about access to
compressed-video signals.

For the codec comparison, CoVisco uses H.265/HEVC input from the original
compressed-video stream without re-encoding. Codec selection is performed
independently within each temporal segment, with an equal number of selected
tokens per segment. The per-segment budget may differ between training and
evaluation configurations, but all segments within a given input use the same
budget. In the matched setting, the total CoVisco codec input budget is
$S K_{\mathrm{codec}}=10{,}368$ at both $224\times224$ and $504\times504$
resolutions. The comparison also includes the reported dense-input settings that retain all sampled frame tokens.
OneVision-Encoder uses 10{,}368 codec-selected tokens from a 64-frame input or
8 uniformly sampled frames at $504\times504$, while SigLIP2 uses 8 uniformly
sampled frames at $512\times512$. These budgets refer to tokens entering the
respective vision encoders; LLM-side abstract-only and abstract-plus-top-$K$
budgets are reported separately in the main experiments.

\end{document}